\documentclass[compactaffiliation]{Interspeech}

\interspeechcameraready

\title{MSDA: Combining Pseudo-labeling and Self-Supervision for Unsupervised Domain Adaptation in ASR}

\author[affiliation={1,2}]{Dimitrios}{Damianos}
\author[affiliation={2}]{Georgios}{Paraskevopoulos}
\author[affiliation={1}]{Alexandros}{Potamianos}

\affiliation{Speech and Language Processing Group}{National Technical University of Athens}{Greece}
\affiliation{Institute for Language and Speech Processing}{Athena Research Center}{Greece}
\email{d.damianos@athenarc.gr, g.paraskevopoulos@athenarc.gr, potam@central.ntua.gr}

\usepackage{comment}
\usepackage{multirow}
\usepackage{subcaption}

\begin{document}

\maketitle

\begin{abstract}

In this work, we investigate the Meta PL unsupervised domain adaptation framework for Automatic Speech Recognition (ASR). We introduce a Multi-Stage Domain Adaptation pipeline (MSDA), a sample-efficient, two-stage adaptation approach that integrates self-supervised learning with semi-supervised techniques. MSDA is designed to enhance the robustness and generalization of ASR models, making them more adaptable to diverse conditions. It is particularly effective for low-resource languages like Greek and in weakly supervised scenarios where labeled data is scarce or noisy. Through extensive experiments, we demonstrate that Meta PL can be applied effectively to ASR tasks, achieving state-of-the-art results, significantly outperforming state-of-the-art methods, and providing more robust solutions for unsupervised domain adaptation in ASR. Our ablations highlight the necessity of utilizing a cascading approach when combining self-supervision with self-training.

    
\end{abstract}

\section{Introduction}
\label{sec:introduction}
Automatic Speech Recognition (ASR) has transformed human-machine interaction, driving technologies such as virtual assistants, automated transcription services, and voice-controlled devices. However, ASR systems often face challenges in maintaining accuracy when exposed to data that differ from their training environment. Factors such as background noise, variations in recording equipment, diverse speaking styles, and regional accents can significantly degrade performance. To address these issues, domain adaptation techniques are crucial. They help close the gap between training and real-world deployment conditions, enhancing the system's robustness in diverse and unfamiliar scenarios. 

Unsupervised Domain Adaptation (UDA) methods are a popular choice to mitigate this problem, as they rely on unlabeled data to achieve adaptation to the target domain, which are easily accessible~\cite{long2015learning,ganin2016domain}. In the context of ASR, UDA has been employed to improve robustness across various recording conditions, such as environmental noise and reverberation. In addition, UDA techniques have been applied to cross-lingual and multilingual adaptation tasks, enhancing performance in low-resource languages~\cite{anoop2021unsupervised} and different dialects~\cite{asami2017domain}.

\begin{figure*}[ht]
\centering
    \includegraphics[scale=0.30]{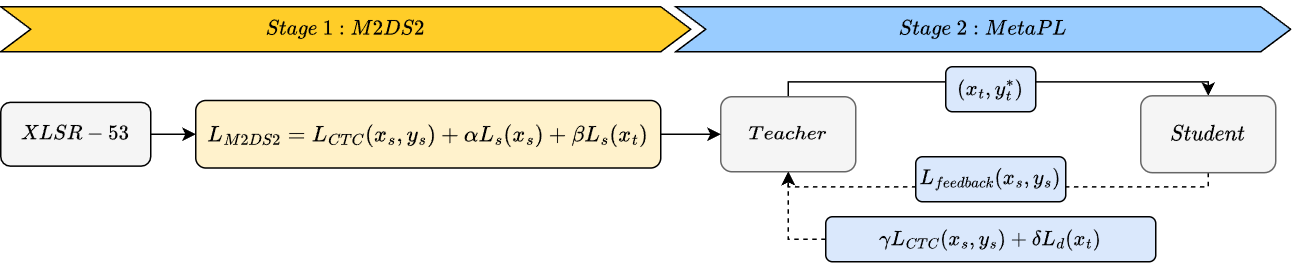}
    \caption{The Multi-Stage Domain Adaptation (MSDA). At Stage 1, we use self-supervision to adapt our pre-trained model to both source $(x_s,y_s)$ and target $(x_t)$ domain, using the loss objective described in Eq.~\eqref{m2ds2_loss}. At Stage 2, we apply an enhanced version of Meta PL to further improve the overall adaptation. In this stage, the teacher provides pseudo-labels $y^*_t$ for student training, and itself is trained using the objective described in Eq.~\eqref{eq:source_teacher_loss}.}
    \label{fig:msda-diagram}
    \vspace{-1em}
\end{figure*}

A prevalent approach to Unsupervised Domain Adaptation (UDA) is teacher-student learning or pseudo-labeling~\cite{scudder1965probability,riloff2003learning}, which reframes the unsupervised task as a supervised one. In this framework, a teacher model trained on the source domain generates pseudo-labels for the target domain, which are then used by a student model for supervised training. However, the teacher’s limited exposure to the target domain often results in erroneous pseudo-labels, leading to the accumulation of errors. To address this, a Confidence Estimation Module (CEM)~\cite{hwang2022large,hwang2022pseudo} filters out unreliable pseudo-labels by estimating the confidence of each prediction, thereby improving training stability and label quality. Additionally, Noisy Student Training (NST)~\cite{park2020improved} enhances domain adaptation by training the student model on heavily augmented target data, such as time masking or speed perturbation, encouraging robustness against domain shifts and improving generalization across domains. Several works propose dynamically updating the teacher model based on the student’s performance to enhance UDA. KAIZEN~\cite{manohar2021kaizen} updates the teacher every $\Delta$ steps using the Exponential Moving Average (EMA) of the student’s parameters, ensuring that the teacher remains a stable yet adaptive guide during training. Meta Pseudo Labels~\cite{pham2021meta} introduces a feedback loop where the teacher is optimized based on the student’s source domain performance, resulting in more accurate pseudo-labels and improved student generalization across domains. Progressive Pseudo-Labeling~\cite{ahmad2024progressive} refines pseudo-labels through an iterative process, where a student trained with hard labels from beam search decoding and a language model subsequently serves as a teacher, progressively improving Word Error Rate (WER). Similarly, kNN-CTC~\cite{zhou2024knn} enhances pre-trained CTC-based ASR models by retrieving fine-grained CTC pseudo-labels from a kNN datastore. Its “skip-blank” strategy effectively addresses alignment challenges in audio-transcript pairs, leading to more accurate predictions and improved UDA performance.

Another widely used approach for domain adaptation is self-supervision, which was first introduced for Natural Language Processing (NLP) tasks~\cite{karouzos2021udalm,gururangan2020don}. This strategy has proven to be an effective and straightforward technique for domain adaptation. The core concept behind self-supervision involves using self-supervised loss during training to adapt large pre-trained models to new domains. Continual Pre-Training (CPT)~\cite{gururangan2020don} has been extensively explored in ASR, involves further pre-training of large pre-trained models on unlabeled data to enhance performance on target domain tasks. Hsu et al.~\cite{hsu2021robust} emphasize the importance of incorporating in-domain unlabeled data during continued pre-training (CPT) to enhance domain adaptation performance, while CASTLE~\cite{zhu2023boosting} demonstrates that integrating CPT with a two-stage pseudo-labeling strategy can substantially reduce error rates in the target domain. Another approach is to integrate self-supervision into a multitask objective for model training, ensuring both domains contribution to the self-supervised task. In UDALM~\cite{karouzos2021udalm}, BERT~\cite{kenton2019bert} is pre-trained on both source and target domains, using the pre-training loss on the target domain during fine-tuning to maintain the adaptation. Similarly,~\cite{paraskevopoulos2023sample} applies self-supervision to ASR domain adaptation, ensuring both domains contribute to the self-supervised task to avoid mode collapse.


Our main contributions are: (a) we demonstrate that the Meta Pseudo Labels framework~\cite{pham2021meta}, originally designed for image recognition tasks,  is a viable adaptation method for ASR. (b) We introduce Multi-Stage Domain Adaptation (MSDA), a novel, sample efficient, two-stage domain adaptation framework that integrates self-supervised techniques from~\cite{paraskevopoulos2023sample} with semi-supervised strategies from~\cite{pham2021meta}, built on the Wav2Vec 2.0 architecture~\cite{baevski2020wav2vec}, achieving state-of-the-art adaptation results. (c) Through extensive experimentation, we demonstrate that the cascading framework is the most effective method to combine self-supervised learning with the Meta PL framework. Our results highlight that MSDA significantly enhances model performance in target domains, demonstrates enhanced sample efficiency, and achieves state-of-the-art results, particularly for low-resource languages and weakly supervised data.

\section{Methodology}
\label{sec:methodology}

Fig.~\ref{fig:msda-diagram} illustrates our proposed approach, which builds on and extends the methodologies presented in~\cite{paraskevopoulos2023sample} and~\cite{pham2021meta}. This approach, termed Multi-Stage Domain Adaptation (MSDA), consists of two sequential adaptation stages: an initial self-supervised stage followed by a semi-supervised stage.

\noindent\textbf{Stage 1:} In this stage, we follow the method proposed in~\cite{paraskevopoulos2023sample} to develop a teacher model trained on both domains in a self-supervised manner, using the following objective:
\begin{equation} \label{m2ds2_loss}
    L_{\rm M2DS2}=L_{\rm CTC}(x_s,y_s)+\alpha L_{\rm s}(x_s)+\beta L_{\rm s}(x_t)
\end{equation}
Here, $L_{\rm CTC}(x_s, y_s)$ is the Connectionist Temporal Classification (CTC) loss~\cite{graves2014towards} applied to source domain samples $(x_s, y_s)$, ensuring accurate alignment between the input speech and corresponding transcriptions. The terms $L_{\rm s}(x_s)$ and $L_{\rm s}(x_t)$ represent contrastive self-supervised losses for the source and target speech data, respectively, following the approach in~\cite{baevski2020wav2vec}. These losses promote the learning of robust, domain-invariant feature representations. The coefficients $\alpha$ and $\beta$ control the contribution of the self-supervised objectives.

\noindent\textbf{Stage 2:} In this stage, we leverage the teacher model trained in Stage 1 to generate pseudo-labels for the target domain. These pseudo-labels are then used to train a student model with the CTC objective. Simultaneously, the teacher model is refined using the following objective: 
\begin{equation} \label{eq:source_teacher_loss}
    L = L_{\rm feedback}(x_s,y_s) + \gamma L_{\rm CTC}(x_s,y_s) + \delta L_{\rm d}(x_t)
\end{equation}
In this formulation, $L_{\rm feedback}$ represents the student's CTC loss on source domain data, providing feedback that guides the teacher to produce higher-quality pseudo-labels. The term $L_{\rm CTC}$ corresponds to the teacher’s own CTC loss in the source domain, ensuring continued alignment with the labeled data. $L_{\rm d}(x_t)$ is a diversity loss, as defined in~\cite{baevski2020wav2vec}, which encourages output diversity to prevent overfitting. The parameters $\gamma$ and $\delta$ are discount coefficients that balance the contribution of each term. 

To effectively integrate self-supervised learning with Meta PL's feedback, we adopt a cascading framework. In Stage 1, we apply SpecAugment~\cite{park2019specaugment} to both domains' audio inputs  to train a more robust teacher. In Stage 2, however,  SpecAugment is only used in student's inputs. We avoid applying it to the teacher's inputs in this stage, as the added noise and augmentation lead to the generation of low-quality pseudo-labels. 

We use Mixed Multi-Domain Self-Supervision (M2DS2) training at the first stage because, as described in~\cite{paraskevopoulos2023sample}, multi-domain self-supervision enables domain adaptation while preventing potential mode collapse. This approach facilitates the generation of higher-quality pseudo-labels in the second stage. Furthermore, incorporating $L_{\rm d}$ into the target domain is essential to prevent mode collapse, while $L_{\rm CTC}$ ensures that the teacher model remains adaptable to the source domain despite adjustments made by $L_{\rm feedback}$.

This two-stage framework not only leverages the strengths of both self-supervised and semi-supervised learning but also enhances domain generalization by maintaining source knowledge while adapting to the target domain. The integration of SpecAugment and diversity loss further improves model robustness, leading to a more stable and reliable adaptation process.

\begin{table*}[htbp]
\centering
\begin{tabular}{c||c|c|c|c|c|c||c}
Adaptation Setting & FT & CPT & M2DS2 & FT-MP & M2DS2-MP & CASTLE & MSDA \\ \hline
LG $\rightarrow$ CV     & 59.57& 66.43&  51.31& 61.76&  55.21&        \underline{51.20}& \textbf{50.18}\\
LG $\rightarrow$ HP     & 62.13& 67.51&  \underline{60.09}&       61.01&          61.98&        \textbf{59.27}&      60.98\\ \hline
CV $\rightarrow$ LG    & 69.55& 71.12&  63.40&       65.89&  \underline{63.33}&        64.71&      \textbf{62.51}\\
CV $\rightarrow$ HP    & 70.72& 73.83&  68.70&       66.64&  \underline{63.18}&        64.51&   \textbf{60.08}\\ \hline
 HP$\rightarrow$CV & 55.90& 59.68& 52.95& 54.50& 50.83& \textbf{46.13}&\underline{47.88}\\
 HP$\rightarrow$LG& 48.65& 52.63& 46.37& 47.03& \underline{46.05}& 48.06&\textbf{45.12}\\\end{tabular}
\caption{MSDA performance compared to our selected baselines, using greedy decoding, evaluated with the WER (\%) metric. A$\rightarrow$B indicates adaptation from A (source) to B (target). }
\label{table:all-results}
\vspace{-1.5em}
\end{table*}

\section{Experimental Setup}
\label{sec:experimental-setup}
\subsection{Pre-trained model:}
For our base model, we utilize XLSR-53~\cite{conneau2021unsupervised} , a state-of-the-art pre-trained speech model developed on the Wav2Vec 2.0~\cite{baevski2020wav2vec} architecture. XLSR-53 stands out for its extensive training on a diverse and extensive corpus, comprising 56,000 hours of speech data across 53 languages. This diverse training set equips XLSR-53 with a robust and comprehensive understanding of various linguistic and acoustic features, making it highly versatile for cross-lingual and multi-lingual applications. We employed the openly available version of XLSR-53\footnote{\scriptsize \url{https://huggingface.co/facebook/wav2vec2-large-xlsr-53}}, which offers a pre-trained implementation ready for fine-tuning and adaptation tasks.

\subsection{Dataset:}
In our experiments we used the following datasets:
    
    \noindent\textbf{Logotypografia}: Logotypografia (LG)~\cite{digilakis2003large} is one of the first corpora for Large Vocabulary Continuous Speech Recognition in Greek. The dataset contains 33,136 newscast utterances, or 72 hours of speech.
    
    \noindent\textbf{Common Voice}: Common Voice (CV)~\cite{ardila2020common} is a multilingual, crowd-sourced dataset developed by Mozilla. For our experiments, we utilize version 9.0 of CV, specifically the Greek corpus, which comprises 12 hours of training speech data.
    
    \noindent\textbf{HParl}: HParl (HP)~\cite{paraskevopoulos2023sample} consists of parliamentary recordings from sessions of the Hellenic Parliament, spanning the years 2018 to 2022, which contains 99 hours of training audio. 
    
    \noindent\textbf{Greek Podcast Corpus:} The Greek Podcast Corpus (GPC)~\cite{paraskevopoulos2024greek} consists of 3,124 hours of weakly-supervised audio sourced from Greek podcasts, spanning 16 diverse categories such as \textit{True Crime} and \textit{Comedy}. Transcriptions were automatically generated using the WhisperX pipeline~\cite{bain2023whisperx}, resulting in pseudo-labeled audio data. For our experiments, we used the GPC-20 subset, which includes 20 hours of audio per category, focusing specifically on the \textit{True Crime}, \textit{Business}, \textit{Comedy}, and \textit{Education} categories.


\subsection{Baselines}
We evaluate the effectiveness of MSDA using the Word Error Rate (WER) metric. All models are tested on the target domain test set. Our approach is compared with the following six baselines:

\noindent\textbf{Finetuning (FT)}: We perform supervised finetuning of XLSR-53 (CTC) using only the source domain data and evaluate on the target domain test set. 

\noindent\textbf{Continual Pre-Training (CPT):} We perform a pretraining phase using the diversity and contrastive losses from Eq.~\eqref{m2ds2_loss} on the target domain train set to create adapted versions of XLSR. These versions are then finetuned on the source domain using the CTC objective and evaluated on the target domain. 

\noindent\textbf{M2DS2:} We train XLSR-53 using the mixed objective of Eq.~\eqref{m2ds2_loss} on both domains and evaluate on the target test data. 

\noindent\textbf{Finetuned-Meta PL (FT-MP):} We employ the Meta PL framework~\cite{pham2021meta}, using the model fine-tuned on the source domain as the teacher, and evaluate on the target domain. 

\noindent\textbf{M2DS2-Meta PL}: We employ the Meta PL framework~\cite{pham2021meta}, using the M2DS2-trained model on both domains as the teacher, and evaluate on the target domain. 

\noindent\textbf{CASTLE:} Finally, we utilize the CASTLE~\cite{zhu2023boosting} framework on our chosen datasets, using the publicly available code\footnote{\scriptsize\url{https://github.com/zhu-han/CASTLE/tree/CASTLE-base}} provided by the authors, and evaluate on the target test set. 

\subsection{Experimental Settings}
For the experiments involving the LG, CV and HP datasets, we apply Eq.~\eqref{m2ds2_loss} and~\eqref{eq:source_teacher_loss} with $\alpha=0.01$, $\beta=0.02$, and $\gamma=\delta=0.0001$. CASTLE~\cite{zhu2023boosting} consists of an initial CPT phase followed by two pseudo-labeling phases, where a 4-gram language model is used to generate higher-quality pseudo-labels via beam search. Since MSDA focuses on the acoustic model, we adapted CASTLE for a fair comparison by directly generating pseudo-labels from the acoustic model using greedy decoding, without any language model assistance. For our weakly-supervised data, the parameters in Eq.~\eqref{m2ds2_loss} and~\eqref{eq:source_teacher_loss} are set to $\alpha=0.01$, $\beta=0.01$, and $\gamma=\delta=0.001$. For Stage 1, we followed the training setup described in~\cite{paraskevopoulos2023sample}. For Stage 2, we used the AdamW optimizer~\cite{loshchilov2017decoupled} with a learning rate of $3 \times 10^{-4}$, no learning rate scheduler, and a maximum of 30 epochs. All experiments were conducted on a single NVIDIA RTX 3090 GPU (24GB) using mixed precision training.



\section{Results}

In Table~\ref{table:all-results}, we compare the performance of MSDA with selected baselines across six adaptation scenarios involving the LG, CV and HP datasets. Meta PL consistently exhibits strong adaptation capabilities, with FT-MP and M2DS2-MP achieving notable WER reductions of 1–4\% and 1–2\%, respectively, compared to the FT and M2DS2 baselines. These results highlight the effectiveness of Meta PL in ASR adaptation. MSDA consistently outperforms all baselines, delivering significant WER improvements. In particular, MSDA achieves WER reductions of 2–10\% compared to FT and outperforms M2DS2 by 1–10\%. Notably, MSDA demonstrates robustness in challenging scenarios where M2DS2 struggles, such as the \textit{HP$\rightarrow$LG} adaptation. Moreover, MSDA surpasses the state-of-the-art CASTLE framework in four out of six adaptation scenarios, achieving absolute WER improvements ranging from 1 to 4\%.

\begin{table}[htbp]
    \resizebox{\columnwidth}{!}{

\begin{tabular}{c||cccc|c}

\begin{tabular}[c]{@{}c@{}}Adaptation\\ Setting\end{tabular} & FT    & M2DS2 & FT-MP & \begin{tabular}[c]{@{}c@{}}M2DS2\\MP\end{tabular} & MSDA  \\ \hline
TC$\rightarrow$Bus                                                        & 48.8  & 45.32 & 43.47 & 42.35    & \textbf{37.98}\\
TC$\rightarrow$Ed                                                         & 62.7  & 57.16 & 58.26 & 56.27    & \textbf{50.3}  \\
TC$\rightarrow$Com                                                        & 61.9  & 68.77 & 57.56 & 57.1     & \textbf{49.21} \\ \hline
Ed$\rightarrow$TC                                                         & 37.28 & 43.23 & \textbf{33.3} & 42.07    & 39.89 \\
Ed$\rightarrow$Bus                                                        & 50.93 & 56.01 & 48.25 & 55.17    & \textbf{47.03} \\
Ed$\rightarrow$Com                                                        & 52.54 & 71.14 & \textbf{50.7}  & 58.06    & 56.16
\end{tabular}

}
\caption{\label{table:gpc-results}
MSDA performance on weakly-supervised data, using WER. We used GPC's TrueCrime (TC) , Education (Ed), Business (Bus) and Comedy (Com) categories as different domains.}
\vspace{-1.5em}
\end{table}

A similar trend is observed in Table~\ref{table:gpc-results}, where we assess MSDA's performance on weakly-supervised data across six adaptation scenarios, comparing it with the FT, M2DS2, FT-MP, and M2DS2-MP baselines. Once again, Meta PL shows consistent adaptation, with FT-MP and M2DS2-MP achieving WER reductions of 1–4\% and 1–2\% compared to FT and M2DS2. MSDA continues to outperform our baselines in most scenarios, achieving significant WER improvements. Specifically, MSDA reduces WER by 3–12\% compared to FT, and outperforms M2DS2 by 7–18\%, even in cases where M2DS2 fails to adapt, such as \textit{Ed$\rightarrow$TC}.

\begin{figure}[t]
    \centering
    \includegraphics[width=\linewidth]{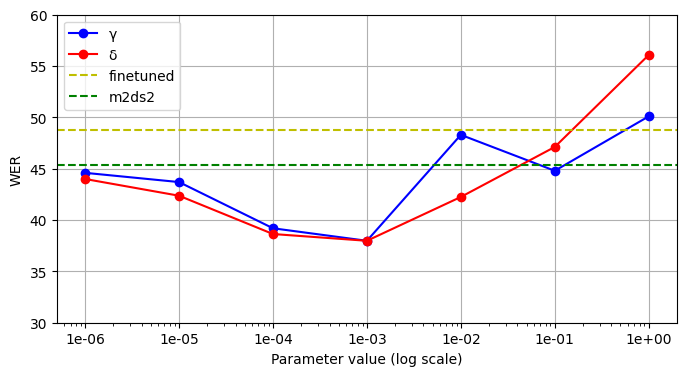}
    \caption{WER comparison between different setting of $\gamma$ (blue) and $\delta$ (red), FT (yellow) and M2DS2 (green) baselines , evaluated in the \textit{TC $\rightarrow$ Ed} adaptation setting.}
    \label{fig:value-comp}
\end{figure}

Fig.~\ref{fig:value-comp} illustrates the effect of $\gamma$ and $\delta$ on MSDA performance for \textit{TrueCrime-Education} adaptation. In two separate experiments, one coefficient is fixed at 0.001 while the other varies from $1$ to $10^{-5}$. MSDA struggles when $\gamma, \delta > 10^{-3}$ due to source overfitting and strong $L_d$ influence, which hinder adaptation. For $\gamma, \delta < 10^{-4}$, performance drops from mode collapse as $L_{CTC}$ and $L_d$ lose effectiveness. The best results are achieved when $\gamma, \delta$ are between $10^{-4}$ and $10^{-3}$.

\begin{figure}[h]
     \centering
     \includegraphics[width=\linewidth]{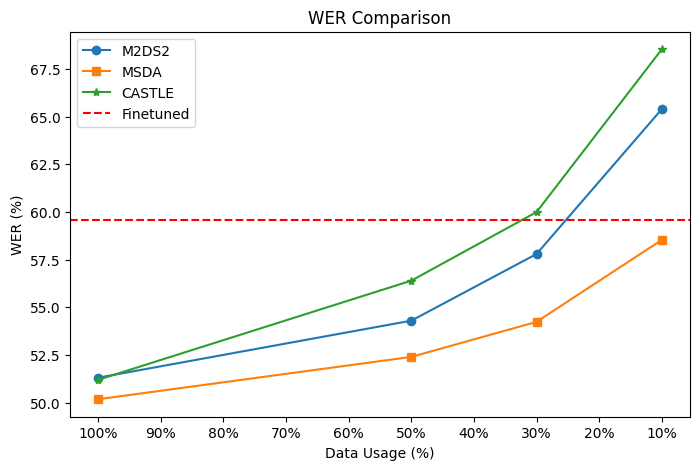}
     \caption{Performance comparison between M2DS2,CASTLE and MSDA for the LG → CV setting, when reducing the amount of available target samples to 50\%, 25\%, and 10\% of the original dataset. FT performance is indicated with the red line.   }
     \label{fig:sample-efficiency}
 \end{figure}

Finally, Fig.~\ref{fig:sample-efficiency} demonstrates the sample efficiency of MSDA in the LG$\rightarrow$CV scenario, compared with M2DS2 and CASTLE, when using 50\%, 25\%, and 10\% of the available target data (corresponding to 6, 3, and 1.2 hours of audio, respectively). The results indicate that MSDA consistently outperforms M2DS2 across all cases, achieving noticeable adaptation even with just 10\% of the available target data. This highlights MSDA's ability to facilitate adaptation even in scenarios where data collection is challenging or impractical.

\section{Meta PL \& Self-Supervision}
\label{sec:failed-attempts}
We explored different ways to integrate self-supervised learning with the Meta PL framework in a non-cascading manner. We experimented with combining self-supervised objectives with student feedback to improve teacher performance, as described in Eq.~\eqref{eq:teach-self-semi}. 
\begin{equation}\label{eq:teach-self-semi}
    L_T = L_{feedback}(x_s,y_s)+L_{s}(x_s)+L_s(x_t)
\end{equation}
In a second approach, we applied self-supervision solely to the target domain for the student model during the pseudo-labeling phase, using Eq.~\eqref{eq:student-self-semi}. 
\begin{equation}\label{eq:student-self-semi}
    L_S = L_{CTC}(x_t,y_t^*)+L_s(x_t)
\end{equation}
Despite these efforts, both approaches failed to achieve successful adaptation, as shown in Table~\ref{table:failed-attempts}, where we experimented with both FT and M2DS2-trained teacher models on the GPC \textit{TC$\rightarrow$Ed} source-target adaptation setting. 

\begin{table}[h]
\resizebox{\columnwidth}{!}{
\begin{tabular}{c|cc|cc}
\multirow{2}{*}{Method}                                               & \multicolumn{2}{c|}{FT}                                    & \multicolumn{2}{c}{M2DS2}                                 \\
                                                                      & \multicolumn{1}{l}{Teacher} & \multicolumn{1}{l|}{Student} & \multicolumn{1}{l}{Teacher} & \multicolumn{1}{l}{Student} \\ \hline
MSDA                                                                  &                             62.7&                              \textbf{58.18}&                             57.16&                             \textbf{50.3}\\ \hline
\begin{tabular}[c]{@{}c@{}}teacher\\ feedback+self super\end{tabular} &                             62.7&                              97&                             57.16&                             95.83\\ \hline
\begin{tabular}[c]{@{}c@{}}student\\ CTC+self super\end{tabular}      &                             62.7&                              72&                             57.16&                            
67.1\end{tabular}
    }
\caption{Alternatives explored during integration on the  TC$\rightarrow$Ed source-target adaptation setting. The ``Teacher" column contains teacher's original WER on target domain, while the ``Student" column contains student's WER after adaptation}
\label{table:failed-attempts}
\vspace{-1.5em}
\end{table}
We found that applying Eq.~\eqref{eq:teach-self-semi} to the teacher model led to the generation of low-quality pseudo-labels, which impaired the student model’s ability to adapt effectively to the target domain. Likewise, using Eq.~\eqref{eq:student-self-semi} on the student model produced misleading feedback for the teacher, ultimately reducing the overall quality of the pseudo-labels.

\section{Conclusions \& Future Work}
\label{sec:conclusions}
In this work, we explored the application of the Meta PL framework~\cite{pham2021meta} in ASR tasks and investigated its integration with self-supervision objectives. We found that Meta PL is an effective adaptation method, providing a straightforward and easily implementable solution. However, our proposed approach, Multi-Stage Domain Adaptation (MSDA), significantly outperforms Meta PL, M2DS2, and CASTLE, as highlighted in Tables~\ref{table:all-results} and~\ref{table:gpc-results}. Furthermore, MSDA is a sample-efficient framework, as shown in Fig.~\ref{fig:sample-efficiency}, making it particularly suitable for scenarios where data collection is limited or infeasible. We also explored alternative ways to integrate self-supervision within the Meta PL framework and concluded that cascading methods show promising results. 
In future work, we plan to investigate the effectiveness of MSDA in additional adaptation settings, such as accent adaptation and cross-lingual adaptation. Furthermore, we aim to evaluate its performance on a broader range of tasks, including image recognition and natural language processing (NLP), to further assess its versatility and robustness across diverse domains. 

\section{Acknowledgements}
This work is co-funded by the European Union’s Digital Europe Programme and by the European Union through the National Strategic Reference Framework (NSRF) 2021–2027, under the Operational Programme “Competitiveness” Grant No 101083565 and Grant No 6002637 respectively, for the project SmartAttica-AtHeNAI DIH (the Attica region - Greek Innovation hub for Artificial Intelligence in Energy and Environment, Supply chain and mobility, Culture and Tourism).

\bibliographystyle{IEEEtran}
\bibliography{mybib}

\end{document}